# A Novel Filter Approach for Band Selection and Classification of Hyperspectral Remotely Sensed Images Using Normalized Mutual Information and Support Vector Machines

Hasna Nhaila, Asma Elmaizi, Elkebir Sarhrouni,
and Ahmed Hammouch

Laboratory LRGE, ENSET, Mohammed V University,
B.P. 6207 Rabat, Morocco
hasnaa.nhaila@gmail.com, asma.elmaizi@gmail.com,
sarhrouni436@yahoo.fr, hammouch_a@yahoo.com

Abstract. Band selection is a great challenging task in the classification of hyperspectral remotely sensed images HSI. This is resulting from its high spectral resolution, the many class outputs and the limited number of training samples. For this purpose, this paper introduces a new filter approach for dimension reduction and classification of hyperspectral images using information theoretic (normalized mutual information) and support vector machines SVM. This method consists to select a minimal subset of the most informative and relevant bands from the input datasets for better classification efficiency. We applied our proposed algorithm on two well-known benchmark datasets gathered by the NASA's AVIRIS sensor over Indiana and Salinas valley in USA. The experimental results were assessed based on different evaluation metrics widely used in this area. The comparison with the state of the art methods proves that our method could produce good performance with reduced number of selected bands in a good timing.

Keywords: Dimension reduction - Hyperspectral images - Band selection
Normalized mutual information – Classification - Support vector machines

## 1 Introduction

Recently, the hyperspectral imagery HSI becomes the principal source of information in many applications such as astronomy, food processing, Mineralogy and specially land cover analysis [1, 2]. The hyperspectral sensors provide more than a hundred of contiguous and regularly spaced bands from visible light to near infrared light of the same observed region. These bands are combined to produce a three dimensional data called a hyperspectral data cube. Thus, an entire reflectance spectrum is captured at each pixel of the scene. This large amount of information increases the discrimination between the different objects of the scene. Unfortunately, we face many challenges in storage, time treatment and especially in the classification schemes due to the many class outputs and the limited number of training samples which is known as the curse of dimensionality [3]. Also the presence of irrelevant and redundant bands complicates the learning algorithms. To overcome these problems, the dimensionality reduction (DR) techniques based on feature selection or extraction become an essential prepro-cessing step that significantly enhances the classification performance. This article will focus on feature selection methods which include filter and wrapper approaches depending on the use of the learning algorithm in the selection process. Our proposed method is a filter approach.

The rest of this paper is organized as follows: Sect. 2 presents the related works of feature selection based methods for dimension reduction of hyperspectral images. In Sect. 3, we describe the proposed approach. Section 4 presents the datasets and discusses the experimental results in comparison with the state of the art methods. Finally, some conclusions of our work are drawn in Sect. 5.

## 2 Related Works

Feature selection approaches have attracted increasing international interest in the last decades. Thus, various methods have been proposed to overcome the HSI classification challenges. The maximal statistical dependency based on mutual information MRMR was used in [4] to select good features for HSI classification. In [5], a greedy opti-mization strategy was applied to select features from HSI data. In their work [6], authors proposed an adaptive clustering for band selection. Additionally, in [7], an unsupervised method for band selection by dominant set extraction DSEBS was pro-posed using the structural correlation. On the other hand, in [8], the Gray Wolf Optimizer GWO was used to reformulate the feature selection as a combinatorial problem based on class separability and accuracy rate by modeling five objective functions.

New methods are still appearing in the literature. In [9], a new method for dimension reduction of Hyperspectral images GLMI was proposed using GLCM fea-tures and mutual information. In [10], authors proposed a semi supervised local Fisher discriminant analysis using pseudo labels samples for dimensionality reduction of HSI. In our work, we propose a new filter approach called NMIBS based on information theoretic, we use the normalized mutual information with the support vector Machi-nes SVM to address the curse of dimensionality problem in HSI classification.

To confirm the effectiveness of our proposed approach, experiments are carried out on the NASA's AVIRIS Indian Pines and Salinas hyperspectral datasets with com-parison to several techniques of band selection and classification of hyperspectral images.

## 3 Methodology

The main aim of this work is to improve the classification performance of hyperspectral images by introducing a new filter approach for band selection. It consists to select the optimal subset of relevant bands and remove the noisy and redundant ones using the normalized mutual information NMI.

According to the general principle of feature selection methods [11], our algorithm comprises four main steps:

- The generation procedure of the candidate bands using sequential feature selection starting with an empty set.
- The evaluation function to judge the goodness of the current subset. In this step, we measure the information and dependence using NMI.
- The stopping criterion to decide when to stop the search. It depends on the number of iterations and the features to be selected.
- The validation procedure to test the effectiveness of the retained subset of bands. In this step, we applied the SVM classifier on tow real world benchmark datasets and compare the obtained results with the state of the art methods.

The remainder of this section gives a brief explanation of the band selection using normalized mutual information, the definition of Support vector machine and presents the complete selection process of the proposed algorithm.

### 3.1 Normalized Mutual Information for Band Selection

Mutual information has been widely studied and successfully applied in hyperspectral remote sensing imagery to select the optimal subset of features [4, 5, 9]. It measures the dependence between two random variables which are, in our case, the ground truth noted GT and each candidate band of the input datasets noted B. In this work, we will use the normalized mutual information given as:

$$NMI(GT, B) = \frac{H(GT) + H(B)}{H(GT, B)} \quad (1)$$

This measure represents the ratio of the entropy of the ground truth GT and each band B on the joint entropy between GT and B. It is higher when we have a good similarity between the bands. Low value means a small similarity and zero value shows that the bands are independent which allows eliminating the noisy bands. The NMI will be used in the generation and evaluation steps of our proposed methodology see the proposed algorithm in the following subsection.

### 3.2 Support Vector Machines

The Support Vector Machines SVM is applied in the validation step of our proposed method to generate the classified maps using the selected bands. It is one of the most useful as supervised classifier in many works related to hyperspectral remotely sensed images applications [12, 13]. Its principle consists to construct an optimal hyperplane of two classes by maximizing the distance between the margins. SVM is adopted in our work since it is able to work with a limited number of training samples. In our experiments, we use the radial basis function RBF as a kernel to map the input data to a higher dimensional space. Three cases of training sets (10%, 25% and 50%) are randomly constructed to train the classifier to show the impact of the training samples size on the classification rate.

## 3.3 Proposed Algorithm

The complete selection process of our proposed methodology is as follows:

```
Input: hyperspectral dataset HSI: Ground truth GT,
dataset bands B.
Output: Group of selected bands and classified maps.
Initialization: number of iterations z, Number of
bands N, Training samples T, Group of selected bands
"empty set" S, Threshold to control redundancy Th, The
number of features to be selected K.
Begin
  Calculation of the normalized mutual information NMI
  between the Ground truth GT and each band B using
  equation 1.
      For b∈ B, compute NMI(GT,b_i)
  Selection process
      Select the first band b_i that maximizes NMI(GT, b_i)
```
$b_i := argmax_{b_i} \text{NMI}(i)$
```
      Set B B\{b_i}; S {b_i} ; GT_est0 := band(b_i)
      Z=0;
      while [S]<k & z<N-1 do
```
$b_i := argmax_{b_i \in (B-S)} \text{NMI}(i)$ & B←B\{$b_i$}

$GT_{est} := (GT_{est} + band(b_i))/2$
```
         Compute NMI(GT, GT_est)
            If NMI> NMI*+Th then
               NMI*:=NMI; GT_est0 := GT_est; S S∪{b_i} ;
            end if
      end while
  Output: S is the set of selected bands.
  Validation using SVM-RBF classifier
      Construct the SVM model using the training set T.
      Train the SVM model.
  Output: Classified maps
  Compute the Confusion matrix to extract evaluation
  metrics.
end
```

In this algorithm, to generate the optimal subset of reduced bands, we initialize the selected bands by the one that have the largest NMI with the ground truth, then an approximated reference map called $GT_{est}$ is built by the average of the last one, firstly named $GT_{est0}$ with the candidate band. The current band is retained if it increases the last value of NMI(GT, $GT_{est}$) used as the evaluation function, otherwise, it will be rejected. The threshold Th is introduced to control the permitted redundancy. The stopping criterion is tested depending on the number of bands to be selected k and the number of iterations z. Finally, the validation step is achieved using the SVM classifier to produce the classified maps as an output of this algorithm. Several evaluations metrics are then calculated based on the confusion matrix for the comparison with various other techniques.

## 4 Experimental Results and Discussion

### 4.1 Datasets Description

In order to evaluate the performances of the proposed approach, the experiments are conducted on two challenging hyperspectral datasets widely used in the literature [14, 15] and freely available in [16]. The first one was captured over Indian pines region in Northwestern Indiana. The second was gathered over Salinas Valley in South California in USA. Both of them are collected by the NASA's Airborne Visible/Infrared Imaging Spectrometer Sensor AVIRIS. Table 1 shows the different characteristics of these datasets. The Color composite and the corresponding ground truth reference with classes are respectively presented in (a) and (b) in Fig. 1.

Table 1. Characteristics of the hyperspectral images used in this work.

|  | Number of bands | Number of classes | Size of the images | Wavelength range | Spatial resolution |
|---|---|---|---|---|---|
| Indian | 224 | 16 | 145  145 | 0.4–2.5 µm | 20 m pixels |
| Salinas | 224 | 16 | 217  512 | 0.4–2.5 µm | 3.7 m pixels |

### 4.2 Parameters Setting and Performance Comparisons

The proposed method is compared with various feature selection methods including Mutual information maximization MIM [17], MRMR [4] and GWO [8]. The SVM classification using all bands without dimension reduction is also included in the comparison.

All the experiments are compiled in the scientific programming language Matlab on a computer with quad-core Duo, 64-b, CPU 2.1 Ghz frequency with 3 GB of RAM. The libsvm package available at [18] was used to get the SVM multiclass classifier with RBF kernel. The proposed algorithm stops when the preferred number of selected bands is achieved. The hyperspectral input datasets are randomly divided into training and testing sets, we consider three cases with ratio of 1:10, 1:4, 1:2.

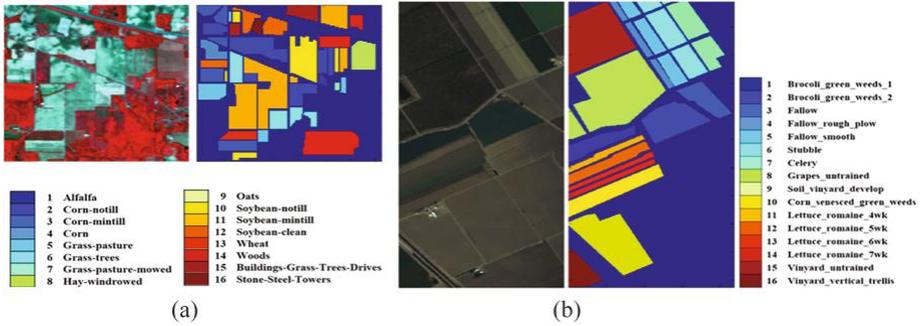

Fig. 1. The Color composite and the corresponding ground truth with class labels for: (a) Indian Pines and (b) Salinas dataset.

### 4.3 Results and Discussion

The experimental results on Indian Pines and Salinas datasets using the proposed approach are presented in this subsection. The classification performances are accessed using two evaluation metrics widely used in the hyperspectral remotely sensed images applications which are: the Average Accuracy AA and the Overall Accuracy OA. The AA measures the average of classification accuracy for all classes; it's calculated as the ratio of the sum of each class accuracy on the number of classes. Whereas, the OA shows the number of correctly predicted pixels over all the test samples. The computational time is also calculated.

Tables 2 and 3 show the Overall accuracy obtained for respectively Indian and Salinas datasets. The first column in each table represents the number of the selected bands. The remainder columns show the obtained OA using different percentage of training samples, we use 10%, 25% and 50%.

Table 2. The Overall Accuracy obtained using the proposed algorithm on Indian Pines datasets for different number of selected bands and training sets.

| Number of selected bands | 10% training | 25% training | 50% training |
|---|---|---|---|
| 10 | 55.2 | 56.72 | 57.33 |
| 20 | 59.65 | 61.28 | 62.76 |
| 30 | 68.23 | 71.61 | 73.98 |
| 40 | 72.06 | 77.29 | 81.93 |
| 50 | 74.00 | 79.65 | 84.84 |
| 60 | 76.41 | 83.63 | 88.63 |
| 70 | 77.60 | 84.56 | 90.24 |
| 80 | 77.83 | 84.55 | 90.74 |
| 90 | 80.90 | 86.98 | 93.90 |
| 100 | 80.83 | 87.25 | 93.48 |
| All bands | 60.74 | 69.42 | 75.72 |

Table 3. The Overall Accuracy obtained using the proposed algorithm on Salinas datasets for different number of selected bands and training sets.

| Number of selected bands | 10% training | 25% training | 50% training |
|---|---|---|---|
| 10 | 80.35 | 81.36 | 81.81 |
| 20 | 88.13 | 88.54 | 88.90 |
| 30 | 89.84 | 90.27 | 90.58 |
| 40 | 90.66 | 91.29 | 91.63 |
| 50 | 91.80 | 92.41 | 92.79 |
| 60 | 92.26 | 92.86 | 93.27 |
| 70 | 92.59 | 93.23 | 93.59 |
| 80 | 92.65 | 93.28 | 93.80 |
| 90 | 92.62 | 93.36 | 93.91 |
| 100 | 92.65 | 93.48 | 94.08 |
| All bands | 87.31 | 88.77 | 90.02 |

From these results, we can make three main remarks:

First, it is obvious that the number of pixels used for training affect the accuracy rate, the OA increases with the size of the training sets in both Indian and Salinas images. For example, with 70 selected bands in Indian Pines scene, we get 77.60%, 84.55% and 90.24% for respectively 10%, 25% and 50% as training sets, see Table 2. For Salinas, we obtain OA of respectively 92.59%, 93.23% and 93.59%, see Table 3. The classified maps obtained for these values are illustrated in Fig. 2 for Indian Pines and in Fig. 3 for Salinas scene.

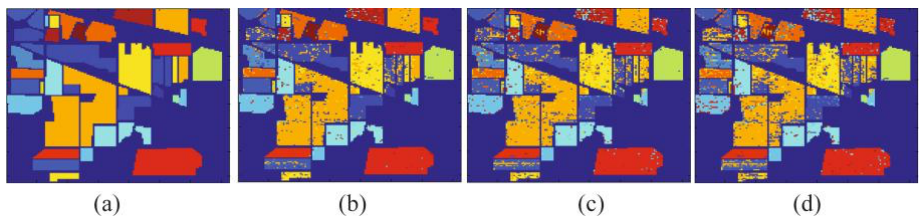

Fig. 2. The ground truth map of Indian Pines dataset (a) and the classified maps using the proposed approach with different training sets: 50% (b), 25% (c) and 10% in (d).

Second, the combination of normalized mutual information and SVM classifier in our proposed methodology produces good classification results even with limited number of training pixels. In the case of 10% as training set, with just 40 selected bands from 224, the OA achieves 72.06% using Indian Pines dataset and 90.66% on Salinas image.

Third, it is clear that the use of a subset of relevant bands gives better classification results than using all bands. In Indian Pines, see Table 2, the OA using all bands is equal to 75,72% whereas it achieves 90.24% with reduced subset of 70 bands.

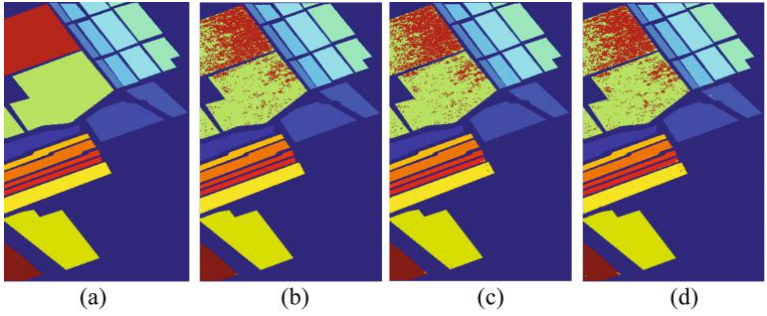

(a)        (b)        (c)        (d)

Fig. 3. The ground truth map of Salinas dataset (a) and the classified maps using the proposed approach with different training sets: 50% (b), 25% (c) and 10% in (d).

In Salinas, we get 87.31% using all bands against 90.66% with just 40 selected bands which confirm the effectiveness of our proposed methodology to select a reduced set of optimal bands and discard the redundant and noisy ones that decrease the classification rate.

In the next experiments, the proposed approach is compared with other methods defined in the literature using only 10% as a training set. The obtained results are presented in Table 4 and evaluated using AA, OA and the running time.

Table 4. The Average Accuracy AA(%), Overall Accuracy OA(%) and computational time (s) obtained by the proposed algorithm in comparison with different methods on Indian Pines and Salinas datasets.

| Methods | Indian Pines dataset | | | Salinas dataset | | |
|---|---|---|---|---|---|---|
| | AA | OA | Time | AA | OA | Time |
| All bands | 42.67 | 60.74 | 42.83 | 91.45 | 87.31 | 397.47 |
| MIM | 56.06 | 73.54 | 12.05 | 93.54 | 88.91 | 126.24 |
| MRMR | 58.70 | 75.70 | 24.87 | 93.56 | 89.67 | 151.55 |
| Gwo-J1 | 67.82 | 71.28 | 170.3 | 94.46 | 89.07 | 1166 |
| Gwo-J2 | 62.57 | 67.44 | 1.7 | 94.68 | 89.25 | 1.05 |
| Gwo-J3 | 64.10 | 70.29 | 0.48 | 94.89 | 89.41 | 5.34 |
| Gwo-J4 | 73.89 | 73.67 | 250 | 97.37 | 95.38 | 1221 |
| Gwo-J5 | 70.43 | 70.65 | 197 | 95.50 | 90.80 | 1198 |
| Proposed NMIBS | 70.41 | 77.90 | 8.77 | 96.47 | 92.54 | 84.67 |

It is seen that our algorithm outperforms the other methods in a good timing. The lower results are obtained by the SVM classification using all bands which confirm the importance of dimension reduction as a preprocessing step of HSI classification to remove the irrelevant bands.

Furthermore, we can see from Table 4 that the running time increases with the size of the used datasets. The classification without dimension reduction needs a significant time compared to the other feature selection methods.

The MIM method outperforms the SVM using all bands but it gives the lower performance in comparison with the other dimension reduction methods (MRMR, GWO and the proposed NMIBS) because it selects bands based only on mutual information maximization without treatment of redundancy between the selected bands.

Gwo-J4 exceeds our methods by 3% but it requires much more time of 250 s against just 8.77 s for our method. In Salinas dataset, the running time of Gwo-J4 is 1121 s against only 84.67 s using our proposed method.

## 5 Conclusion

In this paper, we proposed a new method for band selection to address the curse of dimensionality challenge in hyperspectral images classification. The Normalized Mutual information was adopted to generate and evaluate the selected features using a filter approach. The validation was done using the supervised classifier SVM with RBF kernel.

The experiments were performed on two well-known benchmark datasets collected by the NASA's AVIRIS hyperspectral sensor. Various sets of training and testing samples were randomly constructed to run the proposed algorithm with ratio of 1:10, 1:4 and 1:2. The obtained results were accessed using evaluation metrics widely used in this area.

The comparison with other methods defined in the literature shows the effectiveness of our approach. In overall, we can say that the major advantages of our proposed method is that it is sample, fast and gives a satisfactory results as more complicated methods which we need in the real world applications.